\documentstyle[aclap]{article}

\title{{\small Appears in the Proceedings of the Second Conference on
Empirical Methods in Natural Language Processing, August 1997, Providence, 
RI} \\Distinguishing Word Senses in Untagged Text}

\author{Ted Pedersen \and Rebecca Bruce\\ 
Department of Computer Science and Engineering \\
Southern Methodist University \\
Dallas, TX 75275-0112 \\
\{pedersen,rbruce\}@seas.smu.edu}

\begin{document}
\maketitle

\begin{abstract}
This paper describes an experimental comparison of three unsupervised
learning algorithms that distinguish the sense of an ambiguous word in
untagged text.  The methods described in this paper, McQuitty's
similarity analysis, Ward's minimum--variance method, and the EM
algorithm, assign each instance of an ambiguous word to a known sense
definition based solely on the values of automatically identifiable
features in text.  These methods and feature sets are found to be more
successful in disambiguating nouns rather than adjectives or verbs.
Overall, the most accurate of these procedures is McQuitty's
similarity analysis in combination with a high dimensional feature
set.
\end{abstract}

\section{Introduction}

Statistical methods for natural language processing are often
dependent on the availability of costly knowledge sources such as manually
annotated text or semantic networks. This limits the applicability of such
approaches to domains where this hard to acquire knowledge is already available. 
This paper presents three unsupervised learning algorithms that are able to  
distinguish among the known senses (i.e., as defined in some dictionary)  
of a word, based only on features that can be automatically extracted  
from untagged text.  

The object of unsupervised learning is to determine the class
membership of each observation (i.e. each object to be classified), 
in a sample without using training examples of correct
classifications.  We discuss three algorithms, McQuitty's similarity
analysis \cite{Mcquitty66}, Ward's minimum--variance method
\cite{Ward63} and the EM algorithm \cite{DempsterLR77}, that can be
used to distinguish among the known senses of an ambiguous word
without the aid of disambiguated examples. The EM algorithm produces
maximum likelihood estimates of the parameters of a probabilistic
model, where that model has been specified in advance.  Both Ward's
and McQuitty's methods are agglomerative clustering algorithms that
form classes of unlabeled observations that minimize their respective
distance measures between class members.

The rest of this paper is organized as follows. First, we present
introductions to Ward's and McQuitty's methods (Section 2) and the EM
algorithm (Section 3). We discuss the thirteen words (Section 4) and
the three feature sets (Section 5) used in our experiments. We present
our experimental results (Section 6) and close with a discussion of
related work (Section 7).

\section{Agglomerative Clustering}

In general, clustering methods rely on the assumption that classes 
occupy distinct regions in the feature space. The distance between two
points in a multi--dimensional space can be measured using any of a wide
variety of metrics (see, e.g. \cite{DevijverK82}).  
Observations are grouped in the manner that minimizes the distance between 
the members of each class.

Ward's and McQuitty's method are agglomerative clustering algorithms
that differ primarily in how they compute the distance between
clusters.  All such algorithms begin by placing each observation in a 
unique cluster, i.e. a cluster of one. The two closest clusters are
merged to form a new cluster that replaces the two merged
clusters.  Merging of the two closest clusters continues until only
some specified number of clusters remain.

However, our data does not immediately lend itself to a distance--based
interpretation.  Our features represent part--of--speech (POS) tags,
morphological characteristics, and word co-occurrence; such features
are nominal and their values do not have scale.  Given a POS feature, for 
example, we could choose noun = 1, verb = 2, adjective = 3, and adverb = 4. 
That adverb is represented by a larger number than noun is purely 
coincidental and implies nothing about the relationship between nouns and 
adverbs.

Thus, before we employ either clustering algorithm, we represent our
data sample in terms of a dissimilarity matrix.  Suppose that we have $N$
observations in a sample where each observation has $q$ features.
This data is represented in a $N \times N$ dissimilarity matrix such
that the value in cell $(i,j)$, where $i$ represents the row number
and $j$ represents the column, is equal to the number of features in
observations $i$ and $j$ that do not match.

For example, in Figure \ref{fig:sfv} we have four observations. We
record the values of three nominal features for each observation.
This sample can be represented by the $4 \times 4$ dissimilarity matrix
shown in Figure \ref{fig:shd}.  In the dissimilarity matrix, cells
$(1,2)$ and $(2,1)$ have the value 2, indicating that the first and
second observations in Figure \ref{fig:sfv} have different values for
two of the three features. A value of 0 indicates that observations
$i$ and $j$ are identical.

\begin{figure}
\begin{center}
\begin{tabular}{ccc} 
10 & 2 & 5 \\
1 & 2 & 1 \\
3 & 2 & 5 \\
10 & 2 & 5 
\end{tabular} 
\caption{Matrix of Feature Values}
\label{fig:sfv}
\end{center}
\end{figure}

\begin{figure}
\begin{center}
\begin{tabular}{cccc} 
0 & 2 & 1 & 0 \\
2 & 0 & 2 & 2 \\
1 & 2 & 0 & 1 \\
0 & 2 & 1 & 0 
\end{tabular} 
\caption{Dissimilarity Matrix}
\label{fig:shd}
\end{center}
\end{figure}

When clustering our data, each observation is represented by its
corresponding row (or column) in the dissimilarity matrix.  Using this
representation, observations that fall close together in feature space
are likely to belong to the same class and are grouped together into
clusters.  In this paper, we use Ward's and McQuitty's methods to
form clusters of observations, where each observation is represented
by a row in a dissimilarity matrix.

\subsection{Ward's minimum--variance method}

In Ward's method, the internal variance of a cluster
is the sum of squared distances between each
observation in the cluster and the mean observation for that cluster   
(i.e., the average of all the observations in the cluster). 
At each step in Ward's method, a new cluster, $C_{KL}$, with the smallest 
possible internal variance, is created by merging the two
clusters, $C_K$ and $C_L$, that have the minimum variance between
them. The variance between $C_K$ and $C_L$ is computed
as follows:

\begin{equation}
V_{KL} = \frac{{|| \overline{x}_K - \overline{x}_L||}^2}{\frac{1}{N_K} +
\frac{1}{N_L}}
\end{equation}
where $\overline{x}_K$ is the mean observation for cluster $C_K$,
$N_K$ is the number of observations in $C_K$, and $\overline{x}_L$ and 
$N_L$ are defined similarly for $C_L$.

Implicit in Ward's method is the assumption that the sample comes from a
mixture of normal distributions.  While NLP data is typically not well
characterized by a normal distribution (see, e.g. \cite{Zipf35},
\cite{PedersenKB96}), there is evidence that our data, when represented 
by a dissimilarity matrix, can be adequately characterized by a normal  
distribution.  
However, we will continue to investigate the appropriateness of this
assumption. 

\subsection{McQuitty's similarity analysis}

In McQuitty's method, clusters are based on a simple averaging 
of the feature mismatch counts found in the dissimilarity matrix. 

At each step in McQuitty's method, a new cluster, $C_{KL}$, is formed by 
merging the clusters $C_K$ and $C_L$ that have the fewest number of 
dissimilar features between them. 
The clusters to be merged, $C_K$ and $C_L$, are identified by finding the 
cell $(l,k)$ (or $(k,l)$), where $k \neq l$, that has the minimum value in 
the dissimilarity matrix.  

Once the new cluster $C_{KL}$ is created, the dissimilarity matrix is 
updated to reflect the number of dissimilar features between  
$C_{KL}$ and all other existing clusters. The dissimilarity between
any existing cluster
$C_I$ and $C_{KL}$ is computed as:
\begin{equation}
D_{KL-I} = \frac{D_{KI} + D_{LI}}{2}
\end{equation}
where $D_{KI}$ is the number of dissimilar features between clusters
$C_K$ and $C_I$ and $D_{LI}$ is similarly defined for clusters
$C_L$ and $C_I$. This is simply the average number of mismatches
between each component of the new cluster and the existing cluster.

Unlike Ward's method, McQuitty's method makes no assumptions concerning the 
distribution of the data sample.  

\section{EM Algorithm}

The expectation maximization algorithm \cite{DempsterLR77},
commonly known as the EM algorithm, is an iterative estimation procedure in 
which a problem with missing data is recast to make use of complete data 
estimation techniques. 
In our work, the sense of an ambiguous word is represented 
by a feature whose value is missing. 

In order to use the EM algorithm, the parametric form of the model
representing the data must be known.  In these experiments, we assume
that the model form is the Naive Bayes \cite{DudaH73}. In this model,
all features are conditionally independent given the value of the
classification feature, i.e., the sense of the ambiguous word.  This
assumption is based on the success of the Naive Bayes model when
applied to supervised word--sense disambiguation
(e.g. \cite{GaleCY92}, \cite{LeacockTV93}, \cite{Mooney96},
\cite{PedersenBW97}, \cite{PedersenB97A}).

There are two potential problems when using the EM algorithm.  First,
it is computationally expensive and convergence can be slow for
problems with large numbers of model parameters.  Unfortunately there is 
little to be done in this case other than reducing the
dimensionality of the problem so that fewer parameters are 
estimated.  Second, if
the likelihood function is very irregular it may always
converge to a local maxima and not find the global maximum. 
In this case, an alternative is to use the more computationally 
expensive method of Gibbs Sampling \cite{GemanG84}. 

\subsection{Description}

At the heart of the EM Algorithm lies the Q-function. This 
is the expected value of the log-likelihood function for the complete data  
$D=(Y,S)$, where $Y$ is the observed data and $S$ is the missing sense  
value:
\begin{equation}
Q(\theta^i|\theta) \ \ \ = \ \ \
E[\ln p(Y,S|\theta^i)|\theta,Y)]
\label{eq:em1}
\end{equation}
Here, $\theta$ is the current value of the maximum likelihood
estimates of the model parameters and $\theta^i$ is the improved
estimate that we are seeking; $p(Y,S|\theta^i)$ is the likelihood
of observing the complete data given the improved estimate of the
model parameters. 

When approximating the maximum of the likelihood function, the EM
algorithm starts from a randomly generated initial estimate of $\theta$ 
and then replaces $\theta$ by the $\theta^i$ which maximizes 
$Q(\theta^i|\theta)$. This process is broken down into two steps:
expectation (the E-step), and maximization (the M-step).  The E-step
finds the expected values of the sufficient statistics of the complete
model using the current estimates of the model parameters.  The M-step
makes maximum likelihood estimates of the model parameters using the
sufficient statistics from the E-step.  These steps iterate until the
parameter estimates $\theta$ and $\theta^i$ converge. 

The M-step is usually easy, assuming it is easy for the complete data
problem; the E-step is not necessarily so. However, for decomposable
models, such as the Naive Bayes, the E-step simplifies to the
calculation of the expected counts in the marginal distributions of
interdependent features, where the expectation is with respect to
$\theta$. The M-step simplifies to the calculation of new parameter
estimates from these counts.  Further, these expected counts can be
calculated by multiplying the sample size $N$ by the probability of
the complete data within each marginal distribution given $\theta$ and
the observed data within each marginal $Y_{m}$.  This simplifies to:
\begin{eqnarray*}
{count}^i(S_{m},Y_{m}) = P(S_{m}|Y_{m}) \times  count(Y_{m})
\end{eqnarray*}
where ${count}^i$ is the current estimate of the expected 
count and $P(S_{m}|Y_{m})$ is formulated using $\theta$. 

\subsection{Example} 
For the Naive Bayes model with 3 observable features $A, B,  
C$ and an unobservable classification feature $S$, where $\theta$ = 
$\{P(a,s),P(b,s),P(c,s),P(s)\}$, the E and M-steps are:

\begin{enumerate}
\item {\bf E-step}: The expected values of the sufficient statistics  
are computed as follows:
\begin{eqnarray*}
count^i(s,a) = P(s|a) \times count(a) \\
count^i(s,b) = P(s|b) \times count(b) \\
count^i(s,c) = P(s|c) \times count(c) \\
count^i(s) = \sum_{a,b,c} \{P(s|a,b,c) \times count(a,b,c)\}
\end{eqnarray*}
where:
\begin{eqnarray*}
P(s|a) = \sum_{b,c} P(s|a,b,c)  \\
P(s|a,b,c) = \frac{P(s,a,b,c)}{P(a,b,c)} \\
P(s,a,b,c) = \frac{P(s,a) \times P(s,b) \times P(s,c)}{P(s)^2} \\
P(a,b,c) = \sum_{s} \frac{P(s,a) \times P(s,b) \times P(s,c)}{P(s)^2}
\end{eqnarray*}

\item {\bf M-step}: The sufficient statistics from the E-step are used to
re--estimate the model parameters $\theta^i$: 
\begin{eqnarray*}
P^i(s,a) = \frac{count^i(s,a)}{N} \\
P^i(s,b) = \frac{count^i(s,b)}{N}\\
P^i(s,c) = \frac{count^i(s,c)}{N} \\
P^i(s) = \frac{count^i(s)}{N}
\end{eqnarray*}
\end{enumerate}
where $s,a,b,$ and $c$ denote specific values of $S,A,B,$ and $C$
respectively, and $P(s|b)$ and $P(s|c)$ are defined analogously to
$P(s|a)$. 

\section{Experimental Procedure}

Experiments were conducted to disambiguate 13 different words using 
3 different feature sets. 
In these experiments, each of the 3 unsupervised disambiguation
methods is applied to each of the 13 words using each of the 3 feature
sets; this defines a total of 117 different experiments.  
In addition,
each experiment was repeated 25 times in order to study the variance
introduced by randomly selecting initial parameter estimates, in the
case of the EM algorithm, and randomly selecting among equally distant
groups when clustering using Ward's and McQuitty's methods.

In order to evaluate the  unsupervised learning algorithms we use 
sense--tagged 
text in these experiments. However, this text is only used to
evaluate the accuracy of our methods. The classes discovered by the
unsupervised learning algorithms are mapped to dictionary senses in a
manner that maximizes their agreement with the  sense--tagged text. If the
sense--tagged text were not available, as would often be the case in an 
unsupervised experiment, this mapping would have to be performed manually. 

The words disambiguated and their sense distributions are shown in
Figure \ref{fig:senses}. All data, with the exception of the data for
{\it line}, come from the ACL/DCI Wall Street Journal corpus
\cite{MarcusSM93}.  With the exception of {\it line}, each ambiguous
word is tagged with a single sense defined in the Longman Dictionary
of Contemporary English (LDOCE) \cite{Procter78}. The data for the 12
words tagged using LDOCE senses are described in more detail in
\cite{BruceWP96}.

The {\it line} data comes from both the ACL/DCI WSJ corpus and the
American Printing House for the Blind corpus. Each occurrence of {\it
line} is tagged with a single sense defined in WordNet
\cite{Miller95}. This data is described in more detail in
\cite{LeacockTV93}.

Every experiment utilizes all of the sentences available for each word. 
The number of sentences available per word is shown as ``total count'' 
in Figure \ref{fig:senses}. We have reduced the sense inventory of these
words so that only the two or three most frequent senses are
included in the text being disambiguated.  For several of
the words, there are minority senses that form a very small percentage
(i.e., $<$ 5\%) of the total sample. 
Such minority classes are not yet well handled by unsupervised techniques; 
therefore we do not consider them in this study.

\begin{figure}
\begin{center}
\begin{tabular}{|lr@{}|} \hline
\multicolumn{2}{|c|}{Adjective Senses} \\ 
\hline
\multicolumn{2}{|l|}{{\it chief}: (total count: 1048)} \\
highest in rank: & 86\% \\
most important; main: & 14\%  \\
\hline
\multicolumn{2}{|l|}{{\it common}: (total count: 1060)}\\
as in the phrase `common stock': &  84\% \\
belonging to or shared by 2 or more: & 8\% \\
happening often; usual: & 8\% \\
\hline
\multicolumn{2}{|l|}{{\it last}: (total count: 3004)} \\
on the occasion nearest in the past: & 94\%  \\
after all others: & 6\% \\
\hline
\multicolumn{2}{|l|}{{\it public}: (total count: 715)} \\
concerning people in general: & 68\% \\
concerning the government and people: & 19\% \\
not secret or private: & 13\% \\ 
\hline 
\multicolumn{2}{|c|}{Noun Senses} \\ \hline
\multicolumn{2}{|l|}{{\it bill}: (total count: 1341)} \\
a proposed law under consideration: & 68\% \\
a piece of paper money or treasury bill: & 22\% \\
a list of things bought and their price: & 10\% \\
\hline
\multicolumn{2}{|l|}{{\it concern}: (total count: 1235)} \\
a business; firm:	& 64\% \\
worry; anxiety: & 36\% \\
\hline
\multicolumn{2}{|l|}{{\it drug}: (total count: 1127)} \\
a medicine; used to make medicine: & 57\% \\
a habit-forming substance: & 43\% \\
\hline
\multicolumn{2}{|l|}{{\it interest}: (total count: 2113)} \\
money paid for the use of money:	& 59\% \\
a share in a company or business: & 24\% \\
readiness to give attention:  & 17\% \\
\hline
\multicolumn{2}{|l|}{{\it line}: (total count: 1149)} \\
a wire connecting telephones:	& 37\% \\
a cord; cable:  & 32\% \\
an orderly series: & 30\% \\
\hline
\multicolumn{2}{|c|}{Verb Senses} \\ \hline
\multicolumn{2}{|l|}{{\it agree}: (total count: 1109)} \\
to concede after disagreement: & 74\% \\
to share the same opinion: & 26\% \\
\hline
\multicolumn{2}{|l|}{{\it close}: (total count: 1354)} \\
to (cause to) end: & 77\% \\
to (cause to) stop operation: & 23\% \\
\hline
\multicolumn{2}{|l|}{{\it help}: (total count: 1267)} \\
to enhance - inanimate object: & 78\% \\
to assist - human object: & 22\% \\
\hline
\multicolumn{2}{|l|}{{\it include}: (total count: 1526)} \\
to contain in addition to other parts: & 91\% \\
to be a part of - human subject:  & 9\% \\
\hline
\end{tabular}
\end{center}
\caption{Distribution of Senses}
\label{fig:senses}
\end{figure}

\section{Feature Sets}

We define three different feature sets for use in these 
experiments.  Our objective is to evaluate the effect that different 
types of features have on the accuracy of unsupervised
learning algorithms such as those discussed here. We are particularly
interested  in the impact of the overall dimensionality of the feature
space, and in determining how indicative different feature types are
of word senses. Our feature sets are composed of various combinations of  
the following five types of features.  

\paragraph{Morphology}
The feature M represents the morphology of the ambiguous word.  For
nouns, M is binary indicating singular or plural. For verbs, the
value of M indicates the tense of the verb and can have up to 7
possible values. This feature is not used for adjectives.

\paragraph{Part--of--Speech}

Features of the form $PL_i$ represent the part--of--speech (POS) of the
word $i$ positions to the left of the ambiguous word.  $PR_i$
represents the POS of the word $i$ positions to the right.  In these
experiments, we used 4 POS features, $PL_1$, $PL_2$, $PR_1$, and
$PR_2$ to record the POS of the words 1 and 2 positions to the left
and right of the ambiguous word.  Each POS feature can have one of 5
possible values: noun, verb, adjective, adverb or other.

\paragraph{Co--occurrences}

Features of the form $C_i$ are binary co-occurrence features.  They
indicate the presences or absences of a particular content word in the
same sentence as the ambiguous word.  We use 3 binary co-occurrence
features, $C_1$, $C_2$, and $C_3$ to represent the presences or
absences of each of the three most frequent content words, $C_1$ being
the most frequent content word, $C_2$ the second most frequent and
$C_3$ the third.  Only sentences containing the ambiguous word were
used to establish word frequencies.

Frequency based features like this one contain little information
about low frequency classes.  For words with skewed sense
distribution, it is likely that the most frequent content words will
be associated only with the dominate sense. 

As an example, consider the 3 most frequent content words occurring
in the sentences that contain {\it chief}: {\it officer}, {\it
executive} and {\it president}. {\it Chief}  has a majority class
distribution of 86\% and, not surprisingly, these three content words
are all indicative of the dominate sense which is ``highest in rank''.

The set of content words used in formulating the co--occurrence features 
are shown in Figure \ref{fig:surround}. Note that {\it million} and
{\it  company} occur  frequently. These are not likely to
be indicative of a particular sense but more reflect  the general nature of 
the Wall Street Journal corpus.  

\begin{figure}
\begin{center}
\begin{tabular}{|l|l|l|l|} \hline
\multicolumn{1}{|c|}{word} &   
\multicolumn{1}{|c|}{$C_1$} &   
\multicolumn{1}{|c|}{$C_2$} &   
\multicolumn{1}{|c|}{$C_3$} \\ \hline
chief &   officer & executive & president \\ 
common &   share & million & stock \\ 
last  &   year & week & million \\ 
public  &   offering & million & company \\ 
bill &   treasury & billion & house \\ 
concern  &   million &  company & market\\ 
drug &   fda & company & generic \\ 
interest &   rate & million & company \\ 
line &   he & it & telephone \\ 
agree &   million & company & pay \\ 
close &   trading & exchange & stock \\ 
help &   it & say & he \\ 
include &  million & company & year \\ \hline
\end{tabular} 
\caption{Co--occurrence Features}
\label{fig:surround}
\end{center}
\end{figure}

\paragraph{Unrestricted Collocations}

Features of the form $UL_i$ and $UR_i$ indicate the word occurring in
the position $i$ places to the left or right, respectively, of the
ambiguous word.  
All features of this form have 21 possible values.  Nineteen
correspond to the 19 most frequent words that occur in that fixed
position in all of the sentences that contain the particular ambiguous
word. There is also a value, (none), that indicates when the position
$i$ to the left or right is occupied by a word that is not among the
19 most frequent, and a value, (null), indicating that the position $i$ to 
the left or right falls outside of the sentence boundary.

In these experiments we use 4 unrestricted collocation features,
$UL_2, UL_1, UR_1$, and $UR_2$.  As an example, the values of these
features for {\it concern} are as follows:
\begin{itemize}
\item $UL_2$: and, the, a, of, to, financial, have, because, an, 
's, real, cause, calif., york, u.s., other, mass., german, (null), (none)
\item $UL_1$ : the, services, of, products, banking, 's, pharmaceutical, 
energy, their, expressed, electronics, some, biotechnology, 
aerospace, environmental, such, japanese, gas, investment, (null), (none)
\item $UR_1$: about, said, that, over, 's, in, with, had, are, based, and,
is, has, was, to, for, among, will, did, (null), (none) 
\item $UR_2$: the, said, a, it, in, that, to, n't, is, which, by, 
and, was, has, its, possible, net, but, annual, (null), (none) 
\end{itemize}

\paragraph{Content Collocations}

Features of the form $CL_1$ and $CR_1$ indicate the content word
occurring in the position 1 place to the left or right, respectively,
of the ambiguous word.  The values of these features are defined much
like the unrestricted collocations above, except that these
are restricted to the 19 most frequent content words that occur only
one position to the left or right of the ambiguous word.

To contrast this set of features with the unrestricted collocations,
consider {\it concern} again. The values of the features representing the
19 most frequent content words 1 position to the left and right are as 
follows:
\begin{itemize}
\item $CL_1$: services, products, banking, pharmaceutical, energy, 
expressed, electronics, biotechnology, aerospace, environmental, japanese, 
gas, investment, food, chemical, broadcasting, u.s., industrial, growing, 
(null), (none)
\item $CR_1$: said, had, are, based, has, was, did, owned, were, regarding, 
have, declined, expressed, currently, controlled, bought, announced, 
reported, posted, (null), (none)
\end{itemize}

\paragraph{Feature Sets A, B and C}
The 3 feature sets used in these experiments are designated A, B and C and
are formulated as follows:

\begin{itemize}
\item A: $M, PL_2, PL_1, PR_1, PR_2, C_1, C_2, C_3$ \\
Dimensionality: 5,000 -- 35,000
\item B: $M, UL_2, UL_1, UR_1, UR_2$ \\
Dimensionality: 194,481 -- 1,361,367
\item C: $M, PL_2, PL_1, PR_1, PR_2, CL_1, CR_1$ \\
Dimensionality: 275,625 -- 1,929,375
\end{itemize}

The dimensionality is the number of possible combinations of feature
values and thus the size of the feature space. These values vary since the
number of possible values for M varies with the part--of--speech of the
ambiguous word. The lower number is associated with adjectives and the
higher with verbs. 

To get a feeling for the adequacy of these feature sets,  
we performed supervised learning experiments with the {\it interest} data
using the Naive Bayes model. We disambiguated 3 senses using a 10:1
training--to--test ratio.  The average accuracies for each feature set
over 100 random trials were as follows: A 80.9\%, B 87.7\%, and C 82.7\%. 

The window size, the number of values for the POS features, and 
the number of words considered in the collocation features are kept
deliberately small in order to control the dimensionality of the problem.
In future work, we will expand all of the above types of features and
employ techniques to reduce dimensionality  along  the lines suggested in
\cite{DudaH73} and \cite{GaleCY95}.    

\section{Experimental Results}

Figure \ref{fig:results} shows the average accuracy and standard
deviation of disambiguation over 25 random trials for each combination of 
word, feature set and learning algorithm. 
Those cases where the average accuracy of one algorithm for a
particular feature set is significantly higher than another algorithm,
as judged by the t-test (p=.01), are shown in bold face.  For each
word, the most accurate overall experiment (i.e., algorithm/feature set
combination), and those that are not significantly less accurate are  
underlined.  Also included in Figure \ref{fig:results} is the percentage of 
each sample that is composed of the majority sense.  
This is the accuracy that can be obtained by a {\it majority
classifier}; a simple classifier that assigns each ambiguous word to
the most frequent sense in a sample.  However, bear in mind that in
unsupervised experiments the distribution of senses is not generally
known.

\begin{figure*}[t]
\begin{center}
\begin{tabular}{|@{}l@{}|@{}c@{}|@{}c@{}|@{}c@{}|@{}c@{}||@{}c@{}|@{}c@{}|@{}c@{}||@{}c@{}|@{}c@{}|@{}c@{}|}
\hline
\multicolumn{1}{|l}{} & 
\multicolumn{1}{|l|}{} & 
\multicolumn{3}{|c||}{Feature Set A} &
\multicolumn{3}{|c||}{Feature Set B} &
\multicolumn{3}{|c|}{Feature Set C} \\ \cline{3-11}
&Maj. & McQuitty & Ward & EM & McQuitty & Ward & EM & McQuitty & Ward & EM
\\
\hline
chief  &.861 &    {\bf .844}$\pm$.05  & .721$\pm$.01 &  .729$\pm$.06 & 
{\bf .831}$\pm$.06 & .611$\pm$.01&
.646$\pm$.01 & \underline{{\bf .856}}$\pm$.00 & .673$\pm$.03& 
.697$\pm$.06 \\
common & .842 & {\bf .648}$\pm$.12 & .513$\pm$.08 & .521$\pm$.00 & 
\underline{{\bf .797}}$\pm$.04&
.444$\pm$.04 & .464$\pm$.06 & \underline{{\bf .799}}$\pm$.06 &
.561$\pm$.05 &
.543$\pm$.09 \\
last   & .940 & .791$\pm$.12  & .598$\pm$.09  &  {\bf .903}$\pm$.00 & 
.541$\pm$.11 &
.659$\pm$.03& 
\underline{{\bf .909}}$\pm$.00 & .636$\pm$.07 & .601$\pm$.08 & 
{\bf .874}$\pm$.07 \\
public & .683  & {\bf .560}$\pm$.08 & .450$\pm$.05 &  .473$\pm$.03 & 
{\bf .558}$\pm$.07 & .461$\pm$.03 & .411$\pm$.03 & 
\underline{{\bf .628}}$\pm$.05 &
.488$\pm$.04 & .507$\pm$.03 
\\ \hline 
adjectives & .832 &   {\bf .711}$\pm$.15 & .571$\pm$.12 & .657$\pm$.18 &  
{\bf .682}$\pm$.15 & .544$\pm$.10 &  .608$\pm$.20 & 
\underline{{\bf .730}}$\pm$.11 & .581$\pm$.08  & .655$\pm$.16  
\\ \hline \hline
bill   & .681 & {\bf .669}$\pm$.08 & {\bf .647}$\pm$.11 & .537$\pm$.05 &
\underline{{\bf .753}}$\pm$.05 & .600$\pm$.04& .624$\pm$.08 & 
{\bf .561}$\pm$.10 & .515$\pm$.04 & {\bf .569}$\pm$.04 \\
concern  & .638 &  .629$\pm$.07 & .741$\pm$.04&  
\underline{{\bf .842}}$\pm$.00 & .679$\pm$.04 & .697$\pm$.02 & 
\underline{{\bf .840}}$\pm$.02 & .614$\pm$.08 & {\bf .758}$\pm$.04 & 
{\bf .758}$\pm$.09 \\
drug     & .567  &  .530$\pm$.03 & .557$\pm$.06& 
\underline{{\bf .658}}$\pm$.03 & 
.521$\pm$.01 & .528$\pm$.00& {\bf .551}$\pm$.05 & .573$\pm$.06 &
{\bf .632}$\pm$.06& 
\underline{{\bf .652}}$\pm$.04 \\
interest & .593 & {\bf .601}$\pm$.04 & {\bf .619}$\pm$.04& {\bf
.616}$\pm$.06 
& \underline{{\bf .653}}$\pm$.06 & .552$\pm$.06 & .615$\pm$.05 & 
\underline{{\bf .651}}$\pm$.02 &.615$\pm$.04 & 
\underline{{\bf .649}}$\pm$.09 \\
line	 & .373  &     .420$\pm$.03 & .441$\pm$.03 & {\bf .457}$\pm$.01 & 
.403$\pm$.02 & .428$\pm$.03 &
\underline{{\bf .474}}$\pm$.03  & .410$\pm$.02 &.427$\pm$.02 & 
{\bf .458}$\pm$.01 \\
\hline
nouns    & .570 &  .570$\pm$.10 & .601$\pm$.12& {\underline{\bf
.622}}$\pm$.14
& .602$\pm$.11 & .561$\pm$.10& {\underline {\bf .621}}$\pm$.13 & 
.562$\pm$.10 &.589$\pm$.12 & {\bf .617}$\pm$.12 \\ \hline \hline
agree    &.740  &  {\bf .610}$\pm$.08 & .547$\pm$.03& {\bf .631}$\pm$.08
 & 
{\underline {\bf .678}}$\pm$.08 & .613$\pm$.04& 
{\underline{\bf .683}}$\pm$.14 & 
\underline{\bf{ .685}}$\pm$.07 &
.601$\pm$.00& \underline{{\bf .685}}$\pm$.14 \\
close    & .771  &   {\bf .616}$\pm$.09 & .531$\pm$.02& .560$\pm$.08 & 
{\bf .667}$\pm$.07 &
{\bf .664}$\pm$.00& {\bf .672}$\pm$.06 & {\underline {\bf .720}}$\pm$.11
&.645$\pm$.04 & .648$\pm$.05 \\
help     & .780 &  {\underline  {\bf .713}}$\pm$.05 & .591$\pm$.05 & 
.586$\pm$.05 & {\bf .636}$\pm$.11 & .519$\pm$.01 & .526$\pm$.00 & 
\underline{{\bf .700}}$\pm$.06 & .570$\pm$.03 & .602$\pm$.03 \\
include  &.910 &  {\underline {\bf .880}}$\pm$.06 & .707$\pm$.08&
.725$\pm$.02
& {\bf .767}$\pm$.09 & {\bf .770}$\pm$.06 & {\bf .783}$\pm$.07 & 
{\bf .768}$\pm$.17
&.558$\pm$.04 & .535$\pm$.00  \\ \hline
\hline
verbs  & .800   &  {\underline{\bf .705}}$\pm$.13 &.594$\pm$.08 &
.626$\pm$.09
& 
{\underline{\bf .687}}$\pm$.10 & .642$\pm$.10 & 
{\underline{\bf .666}}$\pm$ 
.12 & {\underline {\bf .718}}$\pm$.11 & .593$\pm$.05& .618$\pm$.09 \\
\hline
\hline
overall & .734 &  {\bf .655}$\pm$.14 & .589$\pm$.11& .634$\pm$.14 & 
{\bf .653}$\pm$.12& .580$\pm$.11 & .631$\pm$.16& 
{\underline {\bf .662}}$\pm$.13 &.588$\pm$.09 & .629$\pm$.13
\\ \hline
\end{tabular} 
\caption{Experimental Results - accuracy $\pm$ standard deviation}
\label{fig:results}
\end{center}
\end{figure*}

Perhaps the most striking aspect of these results is that, across all
experiments, only the nouns are disambiguated with accuracy greater
than that of the majority classifier.  This is at least partially
explained by the fact that, as a class, the nouns have the most
uniform distribution of senses. This point will be elaborated on in
Section \ref{skewed}.  While the choice of feature set impacts
accuracy, overall it is only to a small degree. We return to this
point in Section \ref{features}.  The final result, to be discussed in
Section \ref{algorithm}, is that the differences in the accuracy of
these three algorithms are statistically significant both on average
and for individual words.

\subsection{Distribution of Classes}
\label{skewed}

Extremely skewed distributions pose a challenging learning problem 
since the sample contains precious little information
regarding minority classes.  This makes it difficult to
learn their distributions without prior knowledge.  For unsupervised
approaches, this problem is exacerbated by the difficultly in
distinguishing the characteristics of the minority classes from noise.

In this study, the accuracy of the unsupervised algorithms
was less than that of the majority classifier in every case where 
the percentage of the majority sense exceeded 68\%.  
However, in 
the cases where the performance of these algorithms was less than that of 
the majority classifier, they were often  still providing high accuracy 
disambiguation (e.g., 91\% accuracy for {\it last}).
Clearly, the 
distribution of classes is not the only factor affecting disambiguation 
accuracy;  compare the performance of these algorithms on {\it bill} and 
{\it public} which have roughly the same class distributions. 

It is difficult to quantify the effect of the distribution of
classes on a  learning algorithm particularly when using
naturally occurring data.  In previous unsupervised experiments with {\it 
interest}, using a modified version of Feature Set A, we were able to
achieve an increase of 36 percentage points
over the accuracy of the majority classifier when the 3 classes
were evenly distributed in the sample \cite{PedersenB97B}.  Here, our best  
performance using a larger sample with a natural distribution of senses  
is only an increase of 20 percentage points 
over the accuracy of the majority classifier. 

Because skewed distributions are common in lexical work
\cite{Zipf35}, they are an important consideration in formulating 
disambiguation experiments.  In future work, we will investigate
procedures for feature selection that are more sensitive to minority
classes.  Reliance on frequency based features, as used in this work,
means that the more skewed the sample is, the more likely it is that the
features will be indicative of only the majority class.  

\subsection{Feature Set}
\label{features}

Despite varying the feature sets, the relative accuracy of the three
algorithms remains rather consistent. For 6 of the  13 words there was a 
single algorithm that was always significantly more accurate than the other 
two across all features. 

The EM algorithm was most accurate for {\it last} and {\it line} with  all 
three feature sets.  McQuitty's method was significantly more accurate for 
{\it chief}, {\it common}, {\it public}, and {\it help} regardless of the 
feature set. 

Despite this consistency,  there were some observable trends associated 
with changes in feature set.  For example, McQuitty's method was 
significantly more accurate overall in  combination  with feature set C 
while the EM algorithm was more accurate with Feature Set A, and the 
accuracy of Ward's method was the  least favorable with Feature Set B.

For the nouns, there was no significant difference between Feature Sets A 
and B when using the EM algorithm. For the verbs there was no significant 
difference between the three feature sets when using McQuitty's method. 
The adjectives were significantly more accurate when using McQuitty's 
method and Feature Set C.

One possible explanation for the consistency of results as feature sets
varied is that perhaps the features most indicative of word senses are
included in all the sets due to the selection   methods and the commonality
of feature types.  These common features may be sufficient for the level of 
disambiguation achieved here. This explanation seems more plausible for  
the EM algorithm, where features are weighted, but less so for McQuitty's 
and Ward's which use a representation that does not allow feature 
weighting. 

\subsection{Disambiguation Algorithm}
\label{algorithm}

Based on the average accuracy over part--of--speech categories, the EM
algorithm performs with the highest accuracy for nouns while
McQuitty's method performs most accurately for verbs and
adjectives. This is true regardless of the feature set employed. 

The standard deviations give an indication of the effect of ties on
the clustering algorithms and the effect of the random initialization
on the the EM algorithm. In few cases is the standard deviation very
small. For the clustering algorithms, a high standard deviation
indicates that ties are having some effect on the cluster
analysis. This is undesirable and may point to a need to expand the
feature set in order to reduce ties. For the EM algorithm, a high
standard deviation means that the algorithm is not settling on any
particular maxima.  Results may become more consistent if the number
of parameters that must be estimated was reduced.

Figures \ref{fig:concerna}, \ref{fig:interestb} and \ref{fig:helpc}  
show the confusion matrices associated with the disambiguation of {\it 
concern}, {\it interest}, and {\it help}, using Feature Sets A, B, and C, 
respectively. A confusion matrix shows the number of cases where  the sense 
discovered by the algorithm agrees with the manually assigned sense along 
the main diagonal; disagreements are shown in the rest of the matrix. 

In general, these matrices reveal that both the EM algorithm and
Ward's method are more biased toward balanced distributions of senses
than is McQuitty's method. This may explain the better performance of
McQuitty's method in disambiguating those words with the most skewed
sense distributions, the adjectives and adverbs. It is possible to
adjust the EM algorithm away from this tendency towards discovering
balanced distributions by providing prior knowledge of the expected
sense distribution. This will be explored in future work.

\begin{figure}
\begin{center}
\begin{tabular}{cr|cc|c} 
\multicolumn{2}{c}{} &
\multicolumn{2}{|c|}{Discovered} \\
\multicolumn{1}{c}{} & 
\multicolumn{1}{c|}{Actual} & 
\multicolumn{1}{|c}{worry} & 
\multicolumn{1}{c|}{business} & \\  \hline
&worry & 166 & 281  & 447 \\
&business & 181  & 607 & 788 \\ \hline
&         & 347 & 888 & 1235 \\ \\
\multicolumn{5}{c}{McQuitty - 773 correct} \\ \\
\multicolumn{2}{c}{} &
\multicolumn{2}{|c|}{Discovered} \\
\multicolumn{1}{c}{} & 
\multicolumn{1}{c|}{Actual} & 
\multicolumn{1}{|c}{worry} & 
\multicolumn{1}{c|}{business} & \\  \hline
&worry & 288 & 159  & 447 \\
&business & 155  & 633 & 788 \\ \hline
&         & 443 & 792 & 1235 \\ \\
\multicolumn{5}{c}{Ward - 921 correct} \\ \\
\multicolumn{2}{c}{} &
\multicolumn{2}{|c|}{Discovered} \\
\multicolumn{1}{c}{} & 
\multicolumn{1}{c|}{Actual} & 
\multicolumn{1}{|c}{worry} & 
\multicolumn{1}{c|}{business} & \\  \hline
&worry & 384 & 63  & 447 \\
&business & 132  & 656 & 788 \\ \hline
&         & 516 & 719 & 1235  \\ \\
\multicolumn{5}{c}{EM - 1040 correct} \\
\end{tabular}
\caption{concern - Feature Set A}
\label{fig:concerna}
\end{center}
\end{figure}

\begin{figure}
\begin{center}
\begin{tabular}{cr|ccc|c} 
\multicolumn{2}{c}{} &
\multicolumn{3}{|c|}{Discovered} \\
\multicolumn{1}{c}{} & 
\multicolumn{1}{c|}{Actual} & 
\multicolumn{1}{|c}{attention} & 
\multicolumn{1}{c}{share} & 
\multicolumn{1}{c|}{money} & \\  \hline
&attention & 53 & 6 & 302 & 361 \\
&share & 58 & 187 & 255 & 500 \\
&money & 108 & 4 & 1140 & 1252 \\ \hline
&       & 219 & 197 & 1697 & 2113 \\ \\
\multicolumn{6}{c}{McQuitty - 1380 correct} \\  \\
\multicolumn{2}{c}{} &
\multicolumn{3}{|c|}{Discovered} \\
\multicolumn{1}{c}{} & 
\multicolumn{1}{c|}{Actual} & 
\multicolumn{1}{|c}{attention} & 
\multicolumn{1}{c}{share} & 
\multicolumn{1}{c|}{money} & \\  \hline
&attention & 280 & 3 & 78 & 361 \\
&share & 240 & 197 & 63 & 500 \\
&money & 559 & 0 & 693 & 1252 \\ \hline
&       & 1079 & 200 & 834 & 2113 \\ \\
\multicolumn{6}{c}{Ward - 1170 correct} \\ \\
\multicolumn{2}{c}{} &
\multicolumn{3}{|c|}{Discovered} \\
\multicolumn{1}{c}{} & 
\multicolumn{1}{c|}{Actual} & 
\multicolumn{1}{|c}{attention} & 
\multicolumn{1}{c}{share} & 
\multicolumn{1}{c|}{money} & \\  \hline
&attention & 127 & 230 & 4 & 361 \\
&share & 134 & 364 & 2 & 500 \\
&money & 320 & 124 & 808 & 1252 \\ \hline
&       & 581 & 718 & 814 & 2113 \\ \\
\multicolumn{6}{c}{EM - 1299 correct} \\
\end{tabular}
\caption{interest - Feature Set B}
\label{fig:interestb}
\end{center}
\end{figure}

\begin{figure}
\begin{center}
\begin{tabular}{cr|cc|c} 
\multicolumn{2}{c}{} &
\multicolumn{2}{|c|}{Discovered} \\
\multicolumn{1}{c}{} & 
\multicolumn{1}{c|}{Actual} & 
\multicolumn{1}{|c}{assist} & 
\multicolumn{1}{c|}{enhance} & \\  \hline
&assist & 45 & 234  & 279 \\
&enhance & 146  & 842 & 988 \\ \hline
&         & 191 & 1076 & 1267 \\ \\
\multicolumn{5}{c}{McQuitty - 887 correct} \\ \\
\multicolumn{2}{c}{} &
\multicolumn{2}{|c|}{Discovered} \\
\multicolumn{1}{c}{} & 
\multicolumn{1}{c|}{Actual} & 
\multicolumn{1}{|c}{assist} & 
\multicolumn{1}{c|}{enhance} & \\  \hline
&assist & 88 & 191  & 279 \\
&enhance & 354  & 634 & 988 \\ \hline
&         & 442 & 825 & 1267 \\ \\
\multicolumn{5}{c}{Ward - 722 correct} \\ \\
\multicolumn{2}{c}{} &
\multicolumn{2}{|c|}{Discovered} \\
\multicolumn{1}{c}{} & 
\multicolumn{1}{c|}{Actual} & 
\multicolumn{1}{|c}{assist} & 
\multicolumn{1}{c|}{enhance} & \\  \hline
&assist & 119 & 160  & 279 \\
&enhance & 344  & 644 & 988 \\ \hline
&         & 463 & 804 & 1267 \\ \\
\multicolumn{5}{c}{EM - 763 correct}
\end{tabular} 
\caption{help - Feature Set C}
\label{fig:helpc}
\end{center}
\end{figure}

\section{Related Work}

Word--sense disambiguation has more commonly been cast as a problem in
supervised learning (e.g., \cite{Black88}, \cite{Yarowsky92},
\cite{Yarowsky93}, \cite{LeacockTV93}, \cite{BruceW94B},
\cite{Mooney96}, \cite{NgL96}, \cite{PedersenBW97}, \cite{PedersenB97A}). 
However, all of these methods require that manually sense tagged text be 
available to train the algorithm.  For most domains such text is not 
available and is expensive to create. It seems more reasonable to assume 
that such text will not usually be available and attempt to pursue 
unsupervised approaches that rely only on the features in a text that can 
be automatically identified.

\subsection{Bootstrapping}

Bootstrapping approaches require a small amount of disambiguated text
in order to initialize the unsupervised learning algorithm.  An early
example of such an approach is described in \cite{Hearst91}. A
supervised learning algorithm is trained with a small amount of
manually sense tagged text and applied to a held out test set. Those
examples in the test set that are most confidently disambiguated are
added to the training sample. 

A more recent bootstrapping approach is described in
\cite{Yarowsky95}. This algorithm requires a small number of training
examples to serve as a seed. There are a variety of options discussed
for automatically selecting seeds; one is to identify collocations
that uniquely distinguish between senses. For {\it plant}, the
collocations {\it manufacturing plant} and {\it living plant} make
such a distinction. Based on 106 examples of {\it manufacturing plant}
and 82 examples of {\it living plant} this algorithm is able to
distinguish between two senses of {\it plant} for 7,350 examples with
97 percent accuracy. Experiments with 11 other words using
collocation seeds result in an average accuracy of 96 percent.

While \cite{Yarowsky95} does not discuss distinguishing more than 2 senses 
of a word, there is no immediate reason to doubt that the ``one sense per
collocation'' rule \cite{Yarowsky93} would still hold for a larger
number of senses. In future work we will evaluate using the ``one sense per 
collocation'' rule to seed our various methods.  This may help in dealing 
with very skewed distributions of senses since we currently select
collocations based simply on frequency. 

\subsection{Clustering}

Clustering has most often been applied in natural language processing
as a method for inducing syntactic or semantically related groupings
of words (e.g., \cite{RosenfeldHS69}, \cite{Kiss73}, \cite{RitterK89},
\cite{PereiraTL93}, \cite{Schutze93}, \cite{Resnik95A}). 

An early application of clustering to word--sense disambiguation is
described in \cite{Schutze92}. There words are represented in terms of
the co-occurrence statistics of four letter sequences.  This
representation uses 97 features to characterize a word, where each
feature is a linear combination of letter four-grams formulated by a
singular value decomposition of a 5000 by 5000 matrix of letter
four-gram co-occurrence frequencies.  The weight associated with each
feature reflects all usages of the word in the sample.  A context vector
is formed for each occurrence of an ambiguous word by summing the
vectors of the contextual words (the number of contextual words
considered in the sum is unspecified).  The set of context vectors for
the word to be disambiguated are then clustered, and the clusters are
manually sense tagged.

The features used in this work are complex and difficult to interpret
and it isn't clear that this complexity is required.
\cite{Yarowsky95} compares his method to \cite{Schutze92}
and shows that for four words the former performs significantly better
in distinguishing between two senses.

Other clustering approaches to word--sense disambiguation have been
based on measures of {\it semantic distance} defined with respect to a
semantic network such as WordNet. Measures of semantic distance are
based on the path length between concepts in a network and are used to
group semantically similar concepts (e.g. \cite{LiSM95}).
\cite{Resnik95B} provides an information theoretic definition of
semantic distance based on WordNet.  

\cite{McDonaldPS90} apply another clustering approach to
word--sense disambiguation (also see \cite{WilksFGMPS90}).
They use co-occurrence data gathered from the
machine-readable version of LDOCE to define neighborhoods of
related words.  Conceptually, the neighborhood of a word is a type of
equivalence class.  It is composed of all other words that co-occur
with the designated word a significant number of times in the LDOCE
sense definitions.  These neighborhoods are used 
to increase the number of words in the LDOCE sense definitions, while
still maintaining some measure of lexical cohesion.  The ``expanded''
sense definitions are then compared to the context of an ambiguous
word, and the sense-definition with the greatest number of word
overlaps with the context is selected as correct.  \cite{GuthrieGWA91} propose
that neighborhoods be subject dependent.  They suggest that a word should 
potentially have different neighborhoods corresponding to the different LDOCE 
subject code.  Subject-specific neighborhoods are composed of words having at 
least one sense marked with that subject code.

\subsection{EM algorithm}

The only other application of the EM algorithm to word--sense
disambiguation is described in \cite{GaleCY95}. There the EM
algorithm is used as part of a supervised learning algorithm to
distinguish city names from people's names. A narrow window of
context, one or two words to either side, was found to perform better
than wider windows. The results presented are preliminary but show an
accuracy percentage in the mid-nineties when applied to {\it Dixon}, a
name found to be quite ambiguous.   

It should be noted that the EM algorithm relates to a large
body of work in speech processing.  The Baum--Welch forward--backward 
algorithm \cite{Baum72} is a specialized form of the EM
algorithm that assumes the underlying parametric model is a hidden
Markov model. The Baum--Welch forward--backward algorithm has been used 
extensively in speech recognition (e.g. \cite{LevinsonRS83}, 
\cite{Kupiec92}), \cite{Jelinek90}).     

\section{Conclusions}

Supervised learning approaches to word--sense disambiguation fall
victim to the knowledge acquisition bottleneck. The creation of sense
tagged text sufficient to serve as a training sample is expensive and
time consuming. This bottleneck is eliminated through the use of
unsupervised learning approaches which distinguish the sense of a word
based only on features that can be automatically identified.

In this study, we evaluated the performance of three unsupervised
learning algorithms on the disambiguation of 13 words in naturally
occurring text.  The algorithms are McQuitty's similarity analysis,
Ward's minimum--variance method, and the EM algorithm.  Our findings
show that each of these algorithms is negatively impacted by highly
skewed sense distributions.  Our methods and feature sets were found to be
most successful in disambiguating nouns rather than adjectives or verbs.
Overall, the most successful of our procedures was McQuitty's similarity
analysis in combination with a high dimensional feature set.  In future
work, we will investigate modifications of these algorithms and feature 
set selection that are more effective on highly skewed sense distributions.

\section{Acknowledgments}

This research was supported by the Office of Naval Research
under grant number N00014-95-1-0776. 


\begin{thebibliography}{}

\bibitem[\protect\citename{Baum}1972]{Baum72}
Baum, L.
\newblock 1972.
\newblock An inequality and associated maximization technique in statistical
  estimation for probabilistic functions of a {M}arkov process.
\newblock In O.~Shisha, editor, {\em Inequalities}, volume~3. Academic Press,
  New York, NY, pages 1--8.

\bibitem[\protect\citename{Black}1988]{Black88}
Black, E.
\newblock 1988.
\newblock An experiment in computational discrimination of {E}nglish word
  senses.
\newblock {\em IBM Journal of Research and Development}, 32(2):185--194.

\bibitem[\protect\citename{Bruce and Wiebe}1994]{BruceW94B}
Bruce, R. and J.~Wiebe.
\newblock 1994.
\newblock Word-sense disambiguation using decomposable models.
\newblock In {\em Proceedings of the 32nd Annual Meeting of the Association for
  Computational Linguistics}, pages 139--146.

\bibitem[\protect\citename{Bruce, Wiebe, and Pedersen}1996]{BruceWP96}
Bruce, R., J.~Wiebe, and T.~Pedersen.
\newblock 1996.
\newblock The measure of a model.
\newblock In {\em Proceedings of the Conference on Empirical Methods in Natural
  Language Processing}, pages 101--112.

\bibitem[\protect\citename{Dempster, Laird, and Rubin}1977]{DempsterLR77}
Dempster, A., N.~Laird, and D.~Rubin.
\newblock 1977.
\newblock Maximum likelihood from incomplete data via the {EM} algorithm.
\newblock {\em Journal of the Royal Statistical Society B}, 39:1--38.

\bibitem[\protect\citename{Devijver and Kittler}1982]{DevijverK82}
Devijver, P. and J.~Kittler.
\newblock 1982.
\newblock {\em Pattern Classification: A Statistical Approach}.
\newblock Prentice Hall, Englewood Cliffs, NJ.

\bibitem[\protect\citename{Duda and Hart}1973]{DudaH73}
Duda, R. and P.~Hart.
\newblock 1973.
\newblock {\em Pattern Classification and Scene Analysis}.
\newblock Wiley, New York, NY.

\bibitem[\protect\citename{Gale, Church, and Yarowsky}1992]{GaleCY92}
Gale, W., K.~Church, and D.~Yarowsky.
\newblock 1992.
\newblock A method for disambiguating word senses in a large corpus.
\newblock {\em Computers and the Humanities}, 26:415--439.

\bibitem[\protect\citename{Gale, Church, and Yarowsky}1995]{GaleCY95}
Gale, W., K.~Church, and D.~Yarowsky.
\newblock 1995.
\newblock Discrimination decisions for 100,000 dimensional spaces.
\newblock {\em Journal of Operations Research}, 55:323--344.

\bibitem[\protect\citename{Geman and Geman}1984]{GemanG84}
Geman, S. and D.~Geman.
\newblock 1984.
\newblock Stochastic relaxation, {G}ibbs distributions and the {B}ayesian
  restoration of images.
\newblock {\em IEEE Transactions on Pattern Analysis and Machine Intelligence},
  6:721--741.

\bibitem[\protect\citename{Guthrie \bgroup et al.\egroup }1991]{GuthrieGWA91}
Guthrie, J., L.~Guthrie, Y.~Wilks, and H.~Aidinejad.
\newblock 1991.
\newblock Subject--dependent co--occurrence and word sense disambiguation.
\newblock In {\em Proceedings of the 29th Meeting of the Association for
  Computational Linguistics}, pages 146--152, Berkeley, CA, June.

\bibitem[\protect\citename{Hearst}1991]{Hearst91}
Hearst, M.
\newblock 1991.
\newblock Noun homograph disambiguation using local context in large text
  corpora.
\newblock In {\em Proceedings of the 7th Annual Conference of the UW Centre for
  the New OED and Text Research: Using Corpora}, Oxford.

\bibitem[\protect\citename{Jelinek}1990]{Jelinek90}
Jelinek, F.
\newblock 1990.
\newblock Self--organized language modeling for speech recognition.
\newblock In Waibel and Lee, editors, {\em Readings in Speech Recognition}.
  Morgan Kaufmann, San Mateo, CA.

\bibitem[\protect\citename{Kiss}1973]{Kiss73}
Kiss, G.
\newblock 1973.
\newblock Grammatical word classes: A learning process and its simulation.
\newblock {\em Psychology of Learning and Motivation}, 7:1--41.

\bibitem[\protect\citename{Kupiec}1992]{Kupiec92}
Kupiec, J.
\newblock 1992.
\newblock Robust part-of-speech tagging using a hidden {M}arkov model.
\newblock {\em Computer Speech and Language}, 6:225--243.

\bibitem[\protect\citename{Leacock, Towell, and Voorhees}1993]{LeacockTV93}
Leacock, C., G.~Towell, and E.~Voorhees.
\newblock 1993.
\newblock Corpus-based statistical sense resolution.
\newblock In {\em Proceedings of the ARPA Workshop on Human Language
  Technology}, pages 260--265, March.

\bibitem[\protect\citename{Levinson, Rabiner, and Sondhi}1983]{LevinsonRS83}
Levinson, S., L.~Rabiner, and M.~Sondhi.
\newblock 1983.
\newblock An introduction to the application of the theory of probabilistic
  functions of a {M}arkov process to automatic speech recognition.
\newblock {\em Bell System Technical Journal}, 62:1035--1074.

\bibitem[\protect\citename{Li, Szpakowicz, and Matwin}1995]{LiSM95}
Li, X., S.~Szpakowicz, and S.~Matwin.
\newblock 1995.
\newblock A {W}ord{N}et-based algorithm for word sense disambiguation.
\newblock In {\em Proceedings of the 14th International Joint Conference on
  Artificial Intelligence}, Montreal, August.

\bibitem[\protect\citename{Marcus, Santorini, and
  Marcinkiewicz}1993]{MarcusSM93}
Marcus, M., B.~Santorini, and M.~Marcinkiewicz.
\newblock 1993.
\newblock Building a large annotated corpus of {E}nglish: The {P}enn
  {T}reebank.
\newblock {\em Computational Linguistics}, 19(2):313--330.

\bibitem[\protect\citename{McDonald \bgroup et al.\egroup }1990]{McDonaldPS90}
McDonald, J., T.~Plate, , and R.~Schvaneveldt.
\newblock 1990.
\newblock Using pathfinder to extract semantic information from text.
\newblock In R.~Schvaneveldt, editor, {\em Pathfinder Associative Networks:
  Studies in Knowledge Organization}. Ablex, Norwood, NJ.

\bibitem[\protect\citename{McQuitty}1966]{Mcquitty66}
McQuitty, L.
\newblock 1966.
\newblock Similarity analysis by reciprocal pairs for discrete and continuous
  data.
\newblock {\em Educational and Psychological Measurement}, 26:825--831.

\bibitem[\protect\citename{Miller}1995]{Miller95}
Miller, G.
\newblock 1995.
\newblock Word{N}et: A lexical database.
\newblock {\em Communications of the ACM}, 38(11):39--41, November.

\bibitem[\protect\citename{Mooney}1996]{Mooney96}
Mooney, R.
\newblock 1996.
\newblock Comparative experiments on disambiguating word senses: An
  illustration of the role of bias in machine learning.
\newblock In {\em Proceedings of the Conference on Empirical Methods in Natural
  Language Processing}, pages 82--91, May.

\bibitem[\protect\citename{Ng and Lee}1996]{NgL96}
Ng, H.T. and H.B. Lee.
\newblock 1996.
\newblock Integrating multiple knowledge sources to disambiguate word sense: An
  exemplar-based approach.
\newblock In {\em Proceedings of the 34th Annual Meeting of the Society for
  Computational Linguistics}, pages 40--47.

\bibitem[\protect\citename{Pedersen and Bruce}1997a]{PedersenB97A}
Pedersen, T. and R.~Bruce.
\newblock 1997a.
\newblock A new supervised learning algorithm for word sense disambiguation.
\newblock In {\em Proceedings of the Fourteenth National Conference on
  Artificial Intelligence}, Providence, RI, July.

\bibitem[\protect\citename{Pedersen and Bruce}1997b]{PedersenB97B}
Pedersen, T. and R.~Bruce.
\newblock 1997b.
\newblock Unsupervised text mining.
\newblock Technical Report 97-CSE-9, Southern Methodist University, June.

\bibitem[\protect\citename{Pedersen, Bruce, and Wiebe}1997]{PedersenBW97}
Pedersen, T., R.~Bruce, and J.~Wiebe.
\newblock 1997.
\newblock Sequential model selection for word sense disambiguation.
\newblock In {\em Proceedings of the Fifth Conference on Applied Natural
  Language Processing}, pages 388--395, Washington, DC, April.

\bibitem[\protect\citename{Pedersen, Kayaalp, and Bruce}1996]{PedersenKB96}
Pedersen, T., M.~Kayaalp, and R.~Bruce.
\newblock 1996.
\newblock Significant lexical relationships.
\newblock In {\em Proceedings of the Thirteenth National Conference on
  Artificial Intelligence}, pages 455--460, Portland, OR, August.

\bibitem[\protect\citename{Pereira, Tishby, and Lee}1993]{PereiraTL93}
Pereira, F., N.~Tishby, and L.~Lee.
\newblock 1993.
\newblock Distributional clustering of {E}nglish words.
\newblock In {\em Proceedings of the 31st Annual Meeting of the Association for
  Computational Linguistics}, pages 183--190, Columbus, OH.

\bibitem[\protect\citename{Procter}1978]{Procter78}
Procter, P., editor.
\newblock 1978.
\newblock {\em Longman Dictionary of Contemporary English}.
\newblock Longman Group Ltd., Essex, UK.

\bibitem[\protect\citename{Resnik}1995a]{Resnik95A}
Resnik, P.
\newblock 1995a.
\newblock Disambiguating noun groupings with respect to {W}ord{N}et senses.
\newblock In {\em Proceedings of the Third Workshop on Very Large Corpora},
  MIT, June.

\bibitem[\protect\citename{Resnik}1995b]{Resnik95B}
Resnik, P.
\newblock 1995b.
\newblock Using information content to evaluate semantic similarity in a
  taxonomy.
\newblock In {\em Proceedings of the 14th International Joint Conference on
  Artificial Intelligence}, Montreal, August.

\bibitem[\protect\citename{Ritter and Kohonen}1989]{RitterK89}
Ritter, H. and T.~Kohonen.
\newblock 1989.
\newblock Self-organizing semantic maps.
\newblock {\em Biological Cybernetics}, 62:241--254.

\bibitem[\protect\citename{Rosenfeld, Huang, and Schneider}1969]{RosenfeldHS69}
Rosenfeld, A., H.~Huang, and V.~Schneider.
\newblock 1969.
\newblock An application of cluster detection to text and picture processing.
\newblock {\em IEEE Transactions on Information Theory}, 15:672--681.

\bibitem[\protect\citename{Sch\"utze}1992]{Schutze92}
Sch\"utze, H.
\newblock 1992.
\newblock Dimensions of meaning.
\newblock In {\em Proceedings of Supercomputing '92}, pages 787--796,
  Minneapolis, MN.

\bibitem[\protect\citename{Sch\"utze}1993]{Schutze93}
Sch\"utze, H.
\newblock 1993.
\newblock Word space.
\newblock In S.~Hanson, J.~Cowan, and C.~Giles, editors, {\em Advances in
  Neural Information Processing Systems 5}. Morgan Kaufmann Publishers.

\bibitem[\protect\citename{Ward}1963]{Ward63}
Ward, J.
\newblock 1963.
\newblock Hierarchical grouping to optimize an objective function.
\newblock {\em Journal of the American Statistical Association}, 58:236--244.

\bibitem[\protect\citename{Wilks \bgroup et al.\egroup }1990]{WilksFGMPS90}
Wilks, Y., D.~Fass, C.~Guo, J.~McDonald, T.~Plate, and B.~Slator.
\newblock 1990.
\newblock Providing machine tractable dictionary tools.
\newblock In J.~Pustejovsky, editor, {\em Theoretical and Computational Issues
  in Lexical Semantics}. MIT Press, Cambridge, MA.

\bibitem[\protect\citename{Yarowsky}1992]{Yarowsky92}
Yarowsky, D.
\newblock 1992.
\newblock Word-sense disambiguation using statistical models of {R}oget's
  categories trained on large corpora.
\newblock In {\em Proceedings of the 14th International Conference on
  Computational Linguistics (COLING-92)}, pages 454--460, Nantes, France, July.

\bibitem[\protect\citename{Yarowsky}1993]{Yarowsky93}
Yarowsky, D.
\newblock 1993.
\newblock One sense per collocation.
\newblock In {\em Proceedings of the ARPA Workshop on Human Language
  Technology}, pages 266--271.

\bibitem[\protect\citename{Yarowsky}1995]{Yarowsky95}
Yarowsky, D.
\newblock 1995.
\newblock Unsupervised word sense disambiguation rivaling supervised methods.
\newblock In {\em Proceedings of the 33rd Annual Meeting of the Association for
  Computational Linguistics}, pages 189--196, Cambridge, MA.

\bibitem[\protect\citename{Zipf}1935]{Zipf35}
Zipf, G.
\newblock 1935.
\newblock {\em The Psycho-Biology of Language}.
\newblock Houghton Mifflin, Boston, MA.

\end{thebibliography}

\end{document}